\pdfoutput=1
\documentclass[11pt]{article}

\usepackage[]{EMNLP2022}

\usepackage{times}
\usepackage{latexsym}

\usepackage[T1]{fontenc}
\usepackage[utf8]{inputenc}

\usepackage{microtype}

\usepackage{inconsolata}

\usepackage{multirow,multicol}
\usepackage{graphicx}
\usepackage{amsmath}
\usepackage{cleveref}
\usepackage[nolist]{acronym}

\usepackage{caption}
\usepackage{subcaption}
\usepackage{booktabs}
\usepackage{xcolor}         % colors
\usepackage{colortbl}

\usepackage{graphicx} 
\usepackage{color}
\usepackage{amsmath,amsthm,amssymb}
\usepackage[mathscr]{euscript}

\newcommand{\bali}{\begin{aligned}}
	\newcommand{\eali}{\end{aligned}}

\newcommand{\Ebb}[0]{\ensuremath{\mathbb{E}} }

\newcommand{\Rbb}[0]{\ensuremath{\mathbb{R}} }

\newcommand{\onev}[0]{\ensuremath{{\bf 1}} }

\newcommand\blfootnote[1]{%
  \begingroup
  \renewcommand\thefootnote{}\footnote{#1}%
  \addtocounter{footnote}{-1}%
  \endgroup
}

\title{Pseudo-OOD training for robust language models}

\author{Dhanasekar Sundararaman$^{1*}$, Nikhil Mehta$^{1*}$,  Lawrence Carin$^1$ \\
  $^1$ Duke University \\
  \texttt{\{ds448,nm208\}@duke.edu} \\}
\begin{document}
\maketitle
\begin{abstract}
%The performance of NLP models in downstream tasks has greatly improved due to their increased capacity, resulting from pre-trained models. It is important to note, however, that their out-of-distribution detection performance is often overlooked, which interferes with their ability to make reliable and generalized predictions. The domain independence of NLP models is limited by the fact that they are trained with task-specific data. Our work identifies tokens that may be potential OOD samples through SentBis, a Mahalanobis distance-based framework. In SentBis, OOD samples are identified during training time, and by maximizing the distance between in-distribution (IND) and pseudo-OOD samples, we achieve significant performance improvements on three target benchmarks with minimal post-hoc fine-tuning. Our method achieved an improvement in AUROC of 4\% and an improvement in FPR of 15\% without any usage of external data during training.
%%%%%%%%%%%%%%%%
While pre-trained large-scale deep models have garnered attention as an important topic for many downstream natural language processing (NLP) tasks, such models often make unreliable predictions on out-of-distribution (OOD) inputs. As such, OOD detection is a key component of a reliable machine learning model for any industry-scale application. Common approaches often assume access to additional OOD samples during the training stage, however, outlier distribution is often unknown in advance.  Instead, we propose a post hoc  framework called \textit{POORE - POsthoc pseudo Ood REgularization}, that generates pseudo-OOD samples using in-distribution (IND) data. The model is fine-tuned by introducing a new regularization loss that separates the embeddings of IND and OOD data, which leads to significant gains on the OOD prediction task during testing. We extensively evaluate our framework on three real-world dialogue systems, achieving new state-of-the-art in OOD detection.
\end{abstract}

\section{Introduction}
\blfootnote{$^{*}$The authors contributed equally to this work}
% In the Introduction, motivate using qualitative analysis the following:
% (1) The limitations of Masker-v1 keywords.
% (2) How Mahalanobis distance based keywords overcome the limitations in (1).

Detecting \ac{OOD}~\cite{goodfellow2014explaining, hendrycks2016baseline, yang2021generalized} samples is vital for developing reliable machine learning systems for various industry-scale applications of natural language understanding (NLP)~\citep{shen2019learning, sundararaman2020methods} including intent understanding in conversational dialogues~\citep{zheng2020OODDialog, li2017adversarial}, language translation~\citep{denkowski2011meteor, sundararaman2019syntax}, and text classification~\cite{aggarwal2012survey, sundararaman2022number}. For instance, a language understanding model deployed to support a chat system for medical inquiries should reliably detect if the symptoms reported in a conversation constitute an \ac{OOD} query so that the model may abstain from making incorrect diagnosis~\cite{siedlikowski2021chloe}. %Motivated by such applications, we propose an efficient post hoc fine-tuning framework that improves the reliability of a trained large scale model. 

Although \ac{OOD} detection has attracted a great deal of interest from the research community~\cite{goodfellow2014explaining, hendrycks17baseline, lee2018simple}, these approaches are not specifically designed to leverage the structure of textual inputs. Consequently, commonly used \ac{OOD} approaches often have limited success in real-world NLP applications. Most prior \ac{OOD} methods for NLP systems~\cite{Larson2019AnED, chen2021gold, kamath2020selective} typically assume additional \ac{OOD} data for outlier exposure~\cite{hendrycks2018deep}. However, such methods risk overfitting to the chosen OOD set, while making the assumption that a relevant OOD set is available during the training stage. Other methods~\cite{gangal2020likelihood, li2021k, kamath2020selective} assume training a calibration model, in addition to the classifier, for detecting ~\ac{OOD} inputs. These methods are computationally expensive as they often require re-training the model on the downstream task.

Motivated by the above limitations, we propose a framework called \ac{POORE} that generates pseudo-OOD data using the trained classifier and the \ac{IND} samples. As opposed to methods that use outlier exposure, our framework doesn't rely on any external OOD set. Moreover, \ac{POORE} can be easily applied to already deployed large-scale models trained on a classification task, without requiring to re-train the classifier from scratch. In summary, we make the following contributions:
\begin{enumerate}
    \setlength{\itemsep}{0em}
    \item We propose a Mahalanobis-based context masking scheme for generating pseudo-OOD samples that can be used during the fine-tuning.
    % Selecting keywords based on Mahalanobis distance, which encourages keywords farther away from the IND distribution to be chosen, essentially acting as psuedo-OOD samples.
    \item We introduce a new \ac{POR} loss that maximizes the distance between IND and generated pseudo-OOD samples to improve the OOD detection.
    % Through \ac{POORE}, we introduce a new method \ac{POR}, that maximizes the distance between any IND and an OOD sample, enabling better OOD-resistant models.
    \item Though extensive experiments on the three benchmarks, we show that our approach performs significantly better than existing baselines.
\end{enumerate}

\section{Related Works} \label{related}

\noindent \textbf{\ac{OOD} Detection}. It is a binary classification problem that seeks to identify unfamiliar inputs during inference from in-distribution (\acs{IND}) data observed during training. Standard~\ac{OOD} methods can be divided into two categories. The first category~\cite{lee2018simple, podolskiy2021revisiting, Nalisnick2019DetectingOI, ren2019likelihood} corresponds to approximating a density $p_{IND}(x)$, where density is used as a confidence estimate for binary classification. The second category of approaches~\cite{hendrycks2016baseline, hendrycks17baseline, li2017adversarial, gal2016dropout} use the predictive probability to estimate the confidence scores. In our experiments, we compare against approaches from both the categories. \\

% Several methods have been developed in other disciplines for OOD, but there is major improvement needed in NLP \cite{hendrycks2020pretrained}. 
%In spite of pre-trained models being regarded as better resistant to OOD data, \cite{moon2021masker} empirically demonstrated that they suffer from data bias. 
% GOLD, MASKER, k-Folden
\noindent \textbf{\ac{OOD} Detection in NLP.} There have been several methods developed for \ac{OOD} detection in NLP. \citet{li2021k} proposed using $k$ sub models, where each model is trained with different masked inputs. \citet{kamath2020selective} uses an external ~\ac{OOD} set to train an additional calibration model for \ac{OOD} detection. Most related to our proposed framework is MASKER~\cite{moon2021masker} that leverages \ac{IND} data to generate pseudo-\ac{OOD} samples, and uses self-supervision loss inspired from \citet{devlin2018bert} and predictive entropy regularization for pseudo-\ac{OOD} inputs. We also use BERT self-supervision inspired keyword masking, however, we propose a novel keyword selection criterion. Moreover, we also introduce a novel model regularization loss that directly increases the distance of \ac{IND} and pseudo-\ac{OOD} samples.
 % , each with one masked label and $k-1$ visible labels. Visible labels are optimized with cross-entropy, while the masked one is optimized with the KL divergence loss. \cite{kamath2020selective} uses an additional calibrator model to train the main question answering (QA) model to refrain from answering when there is a high likelihood of an error. 

\section{Preliminaries and Notations} 
\label{methods}

We consider a deep learning model $g \circ f(x)$ composed of an encoder $f: \mathcal{X} \rightarrow \mathcal{F}$ and a classifier $g$ that maps $f(x)$ to the output space, where $x \in \mathcal{X}$ corresponds to natural sentences composed of a sequence of tokens $v_i \in \mathcal{V}$, $i.e.$ $x = [v_1, \dots, v_T]$, $T$ is the length of the sequence, and $\mathcal{V}$ is the token vocabulary. For a downstream classification task, the class prediction is defined as $p(y|x) = \text{softmax}\left(g \left( f\left((x\right) \right)\right)$. \\
% We will assume that both the models $g$ and $f$ are trained on a downstream classification task in a standard supervised setting, where $p(y|x) = \text{softmax}\left(g \left( f\left((x\right) \right)\right)$. \\

\noindent \textbf{Architecture.} In this work, we construct $f$ using the bi-directional Transformer architecture \cite{vaswani2017attention}. Specifically, we use the encoder architecture proposed in \citet{devlin2018bert} such that $f(x)$ is the final hidden representation of the CLS token. We use a two-layer multi-layer perceptron (MLP) as the classifier $g$. \\
% The pre-trained model is constructed using the bi-directional Transformer architecture \cite{vaswani2017attention}, of which the encoder part \cite{devlin2018bert} alone is used for tasks such as classification.  The encoder $f$ has components including sentence-level as well as token-level classifiers. In the baseline GOLD model, the hidden representations of sentences are passed to a classifier module $g$ that generates logits to measure performance using one of the many evaluation methods. With our approach, we employ both sentence-level and token-level classifiers to perform intent classification, as well as self-supervision, and OOD regularization through keyword selection and context masking explained in \ref{maskingood} . \\

% \subsection{Out-of-Distribution Detection}
% \noindent \textbf{\ac{OOD} Detection}. It is a binary classification problem that seeks to identify unfamiliar inputs during inference from in-distribution (\acs{IND}) data observed during training. \ac{OOD} methods typically learn a confidence estimator that outputs a score $s(x) \in \Rbb$ such that $s(x^{ind}) > s(x^{ood})$, where $x^{ind}$ and $x^{ood}$  are sampled from \acs{IND} distribution ($\mathcal{D}_{IND}$) and \acs{OOD} distribution ($\mathcal{D}_{OOD}$) respectively.\\

\noindent \textbf{Mahalanobis OOD Scoring.} \ac{OOD} methods typically learn a confidence estimator that outputs a score $s(x) \in \Rbb$ such that $s(x^{ind}) > s(x^{ood})$, where $x^{ind}$ and $x^{ood}$  are sampled from \acs{IND} distribution ($\mathcal{D}_{IND}$) and \acs{OOD} distribution ($\mathcal{D}_{OOD}$) respectively. \citet{lee2018simple} proposed using Mahalanobis distance estimator for OOD detection that uses pre-trained features of the softmax neural classifier. Namely, given feature of a test sample $\phi(x)$, the mahalanobis score $s_{M}(x)$ is computed as follows
% \begin{small}
% \begin{adjustbox}{max width=\columnwidth}
% \parbox{\linewidth}{%
\begin{align}
    d(x,c) &= \left(\phi(x) - \hat{\mu}_c\right)^T \; \hat{\Sigma}^{-1} \left(\phi(x) - \hat{\mu}_c\right)\\
    s_{M}(x) &= - \; \min_{c}  d(x,c)
\end{align}
% }
% \end{adjustbox}
% \end{small}
where $\phi$ is an intermediate layer of the neural classifier and $c$ denotes the class. The parameters of the estimator $\{\hat{\mu_c}, \hat{\Sigma}\}$ denote the class-conditional mean and the tied covariance of the IND features.
 
\section{Post hoc Pseudo-OOD Regularization}
% to improve the robustness of the model towards $\ac{OOD}$ samples
In this section, we describe our framework called \acf{POORE}, which uses pseudo-OOD samples for fine-tuning a pre-trained classifier. We first describe our masking-based approach to generate pseudo-OOD samples from the \ac{IND} samples available during training. These generated pseudo-OOD samples are used to regularize the encoder during post-hoc training of a pre-trained classifier, which leads to improved robustness of the model towards $\ac{OOD}$ samples.

\subsection{Masking for Pseudo-OOD Generation}
\label{maskingood}
We perform context masking of \ac{IND} samples for generating pseudo OOD samples. To generate context-masked pseudo \ac{OOD} samples, we first identify a set of tokens $v \in \mathcal{K} \subset \mathcal{V}$ that have high attention scores and consequently, a higher influence in model predictions. Given the set of keywords, we perform random masking of non-keywords in a given IND sample $x$ to generate a pseudo \ac{OOD} sample $\tilde{x}$. \\

\noindent \textbf{Keyword Selection.} We follow the attention-based keyword identification method proposed in \citet{moon2021masker}. The token importance is measured using average model attention values computed in the final layer of the pre-trained transformer encoder. 
%More precisely, the importance score $I(v)$ for each token $v \in \mathcal{V}$ is calculated as
% \begin{align}
%     I(v) &= \frac{1}{n_{v}} \sum_{x \in \mathcal{D}_{IND}} \sum_{i=1}^{i=T} \onev_{v_i = v} \cdot a_i \label{eq:masker_keywords}
% \end{align}
% where $a_i \in [0,1]$ is the attention value for $i^{th}$ token in the last self-attention layer in $f$. The keyword set $\mathcal{K}$ is formed by selecting top $M$ tokens based on the token importance score $m(x)$. 
While this approach generates context-deprived inputs, the identified tokens are uniformly selected from all the IND samples in the training data. Instead, we propose a novel weighting criterion for keyword selection, such that a higher weight is given to the tokens belonging to the training inputs that have a higher distance from the overall IND distribution determined by $\{ \hat{\mu}_c, \Sigma \}$. This encourages the selection of keywords that belong to IND inputs which are far from the estimated \ac{IND} distribution. Specifically, we propose the importance score criterion as follows: 

\begin{small}
\begin{align}
    I_M(v) &= \frac{1}{n_{v}} \sum_{x \in \mathcal{D}_{IND}} \left(\sum_{i=1}^{i=T} \onev_{v_i = v} \cdot a_i\right) \cdot \hat{s}_M(x), \label{eq:masker_maha_keywords}\\
    \text{where} \quad &\hat{s}_M(x) = \frac{s_{max} - s_M(x)}{s_{max} - s_{min}},\\
    &s_{min} = \min_{x\in\mathcal{D}_{IND}}s_M(x'),\\
    &s_{max} = \max_{x\in\mathcal{D}_{IND}}s_M(x').
\end{align}
\end{small}
where $a_i \in [0,1]$ is the attention value for $i^{th}$ token in the last self-attention layer in $f$. The keyword set $\mathcal{K}$ is formed by selecting top $M$ tokens based on the token importance score $I_m(v)$. Note that in (\ref{eq:masker_maha_keywords}), the tokens in the \ac{IND} samples having a lower mahalanobis score ($i.e.$ a higher mahalanobis distance) will be more likely to be selected as a keyword.\\

\noindent \textbf{Context Masking.} 
Given the set of keywords $\mathcal{K}$, pseudo OOD samples can be generated by randomly masking the context. The context in an input $x$ refers to the non-keyword tokens $v \notin \mathcal{K}$. We randomly mask the context tokens to generate a pseudo-OOD input $\tilde{x}$, which we use for regularization to improve the model reliability towards \ac{OOD} inputs. More specifically, we create $\tilde{x} = [\tilde{v}_1, \dots, \tilde{v}_T]$ as follows
\begin{align}
    u &\sim \text{Uniform}(0,1) \\
    \tilde{v}_i &= 
        \begin{cases}
            \text{MASK} & \text{if $p_{mask} \leq u$} \\
            v_i    & \text{otherwise}
        \end{cases}
\end{align}
where MASK is the masking token, $p_{mask}$ is the masking probability and $v_i$ is the $i^{th}$ token in the corresponding IND sample $x$.

\subsection{Pseudo-OOD Regularization}
\label{sec:euclidean}

To increase the Mahalanobis distance of \ac{OOD} inputs relative to their \ac{IND} counterparts, we propose \acf{POR} loss. The \ac{POR} regularization maximizes the distance between the IND sample and its corresponding pseudo-OOD sample. The \ac{POR} loss $L_{POR}$ is defined as
\begin{align}
    \mathcal{L}_{POR} &= - \Ebb_{x\sim\mathcal{D}_{IND}} || f(x) - f(\tilde{x})||_2
\end{align}
where $\tilde{x}$ the context-masked version of $x$.\\

\noindent \textbf{Post hoc Training.} The total loss used during post hoc fine-tuning is defined as
\begin{small}
    \begin{align}
        \mathcal{L} &= \mathcal{L}_{CE} + \lambda_{SKL} \cdot \mathcal{L}_{SKL} + \lambda_{POR} \cdot \mathcal{L}_{POR} \label{eq:posthoc_training}
    \end{align}
\end{small}
\noindent where $\mathcal{L}_{CE}$ is the standard cross-entropy (CE) loss, and $\mathcal{L}_{SKL}$ is the self-supervised keyword loss (SKL) proposed in \citet{devlin2018bert}. The SKL loss has been found to improve the generalization of the model by avoiding overfitting of the model to certain tokens in the training data~\cite{moon2021masker}. The post hoc training process finetunes the model using (\ref{eq:posthoc_training}). Note that the post hoc fine-tuning is carried on a model previously trained on the downstream task using the standard $\mathcal{L}_{CE}$ loss.

\begin{table*}
\small
\centering
\resizebox{0.9\textwidth}{!}{%
\begin{tabular}{l||c|c|c|c|c|c}
\toprule
Methods & \multicolumn{2}{c}{STAR} & \multicolumn{2}{|c}{FLOW}  & \multicolumn{2}{|c}{ROSTD} \\ \midrule
& AUROC $\uparrow$ & FPR@90 $\downarrow$  & AUROC $\uparrow$ & FPR@90 $\downarrow$ & AUROC $\uparrow$ & FPR@90 $\downarrow$ \\ \midrule
Maxprob~\citep{hendrycks17baseline}     & 68.27 & 77.18   & 61.10  & 84.23 & 91.49 & 54.30     \\ %\hline
% ODIN~\cite{liang2018enhancing} & 68.27 &  77.18   & 61.10  &  84.23   & 91.49 &  54.30     \\ % \hline
Dropout~\citep{gal2016dropout}     & 52.77  & 100.0        & 51.86 & 100.0 & 55.25 & 100.0         \\ % \hline
Entropy~\citep{lewis1994sequential} & 70.29 &  77.84   & 62.02 &  79.45   & 91.86    & 53.83    \\ % \hline
Gradient Embed~ & 67.61  & 80.80    & 71.21    & 70.25   & 98.53 &  2.58    \\ % \hline
BERT Embed~\citep{podolskiy2021revisiting} & 71.96    & 73.56   & 61.16    & 87.26   & 98.88 &  2.27    \\ % \hline
Mahalanobis~\citep{lee2018simple} & 76.89    & 65.25   & 73.13 & \textbf{63.84}   & 99.45 &  1.00      \\ % 
% \arrayrulecolor{gray}\cmidrule{1-7}\arrayrulecolor{black}
MASKER~\citep{moon2021masker}     & 71.54 &  72.82   & 68.16  & 67.52   & 86.95 & 54.26 \\ %
\arrayrulecolor{gray}\cmidrule{1-7}\arrayrulecolor{black}
MASKER-Maha (Enhanced) & 79.38    & 59.97   & 72.99 & 65.23   & 99.41 &  1.15\\ %
POORE (Ours)     & \textbf{81.11} & \textbf{48.30} & \textbf{74.08} & 69.26 & \textbf{99.51} & \textbf{0.97}    \\ % \hline
\bottomrule
\end{tabular}%
}
\caption{AUROC and FPR@90 of Baseline and POORE on the three target benchmarks. MASKER uses Maxprob for inference, MASKER-Maha and POORE use use Mahalanobis for OOD detection during inference.}
\label{mainresults}
\end{table*}

\begin{figure*}[!th]
\centering
\begin{subfigure}{.45\columnwidth}
  \centering
  \includegraphics[width=\linewidth]{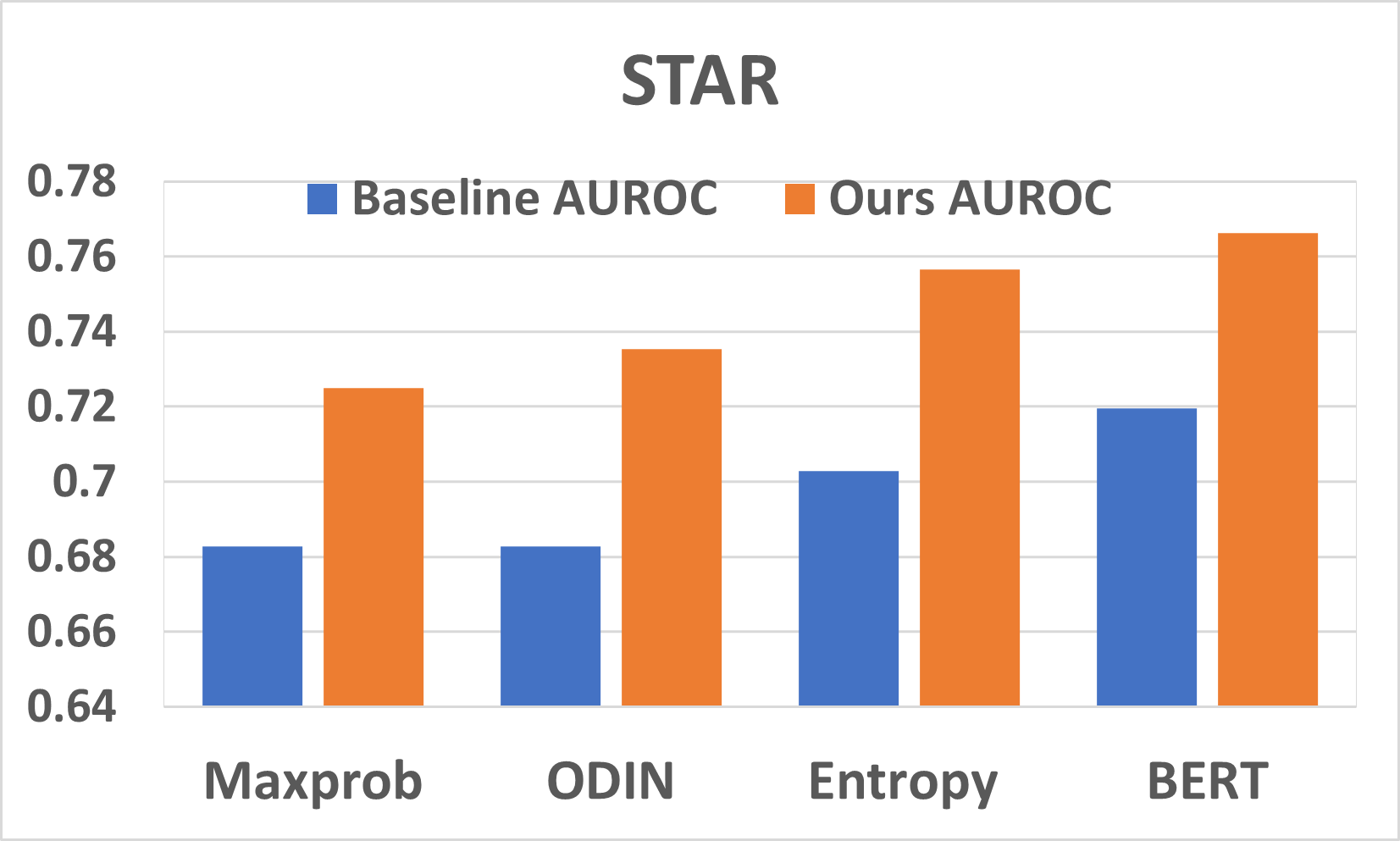}
  \caption{AUROC for STAR}
  \label{fig:sub1a}
\end{subfigure}%
\hfill
\begin{subfigure}{.45\columnwidth}
  \centering
  \includegraphics[width=\linewidth]{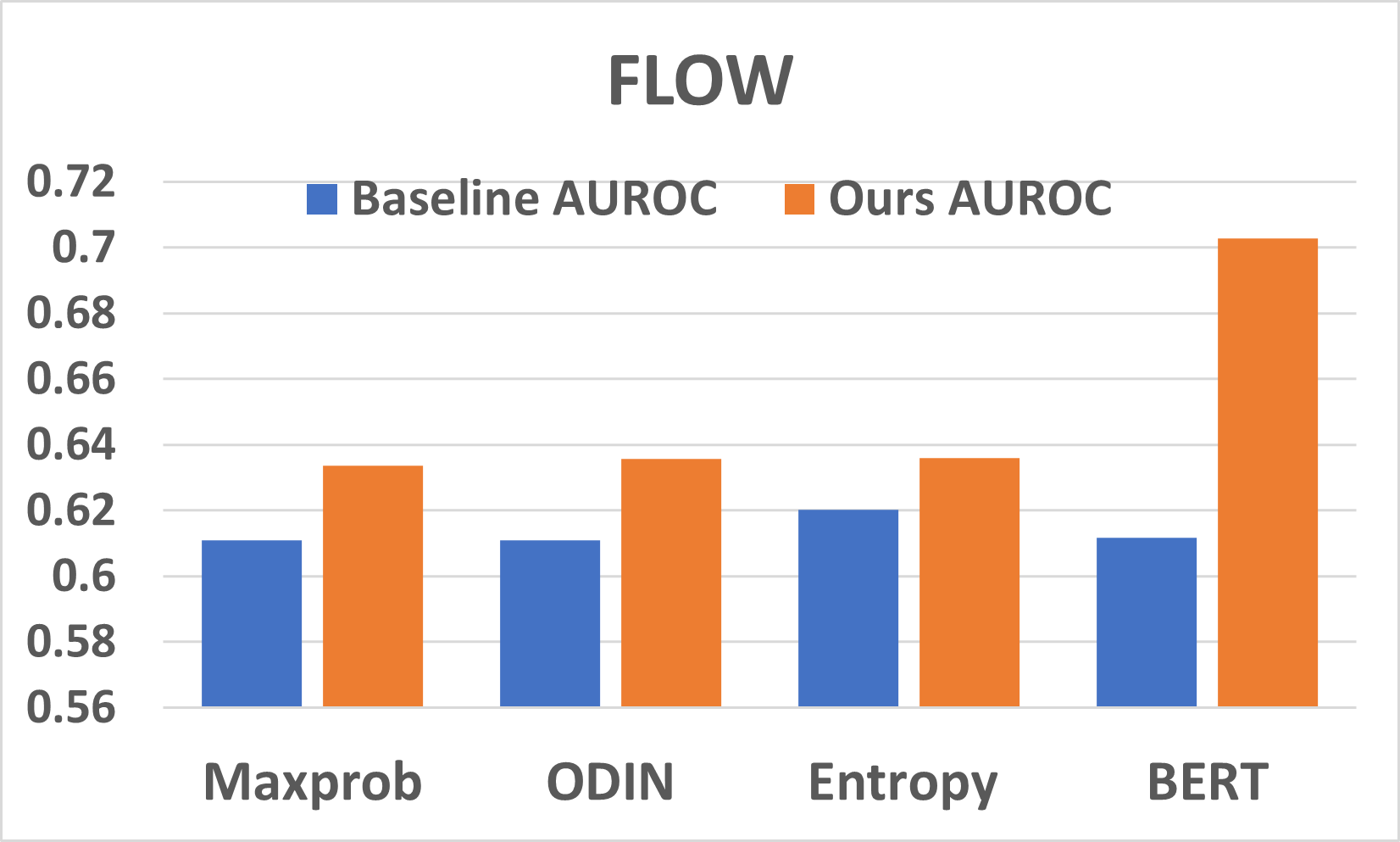}
  \caption{AUROC for FLOW}
  \label{fig:sub1b}
\end{subfigure}
\hfill
\begin{subfigure}{.45\columnwidth}
  \centering
  \includegraphics[width=\linewidth]{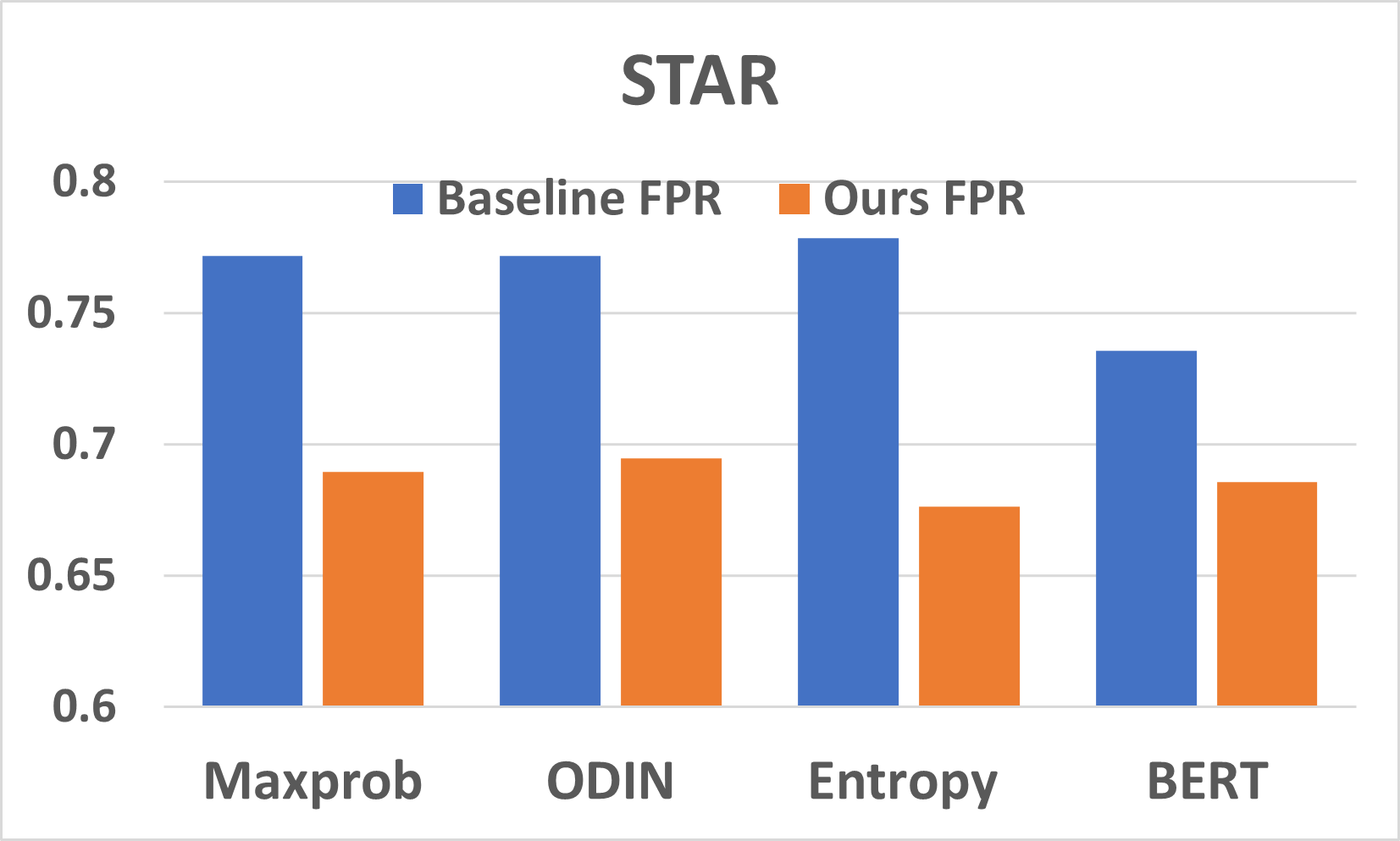}
  \caption{FPR@90 for STAR}
  \label{fig:sub2a}
\end{subfigure}%
\hfill
\begin{subfigure}{.45\columnwidth}
  \centering
  \includegraphics[width=\linewidth]{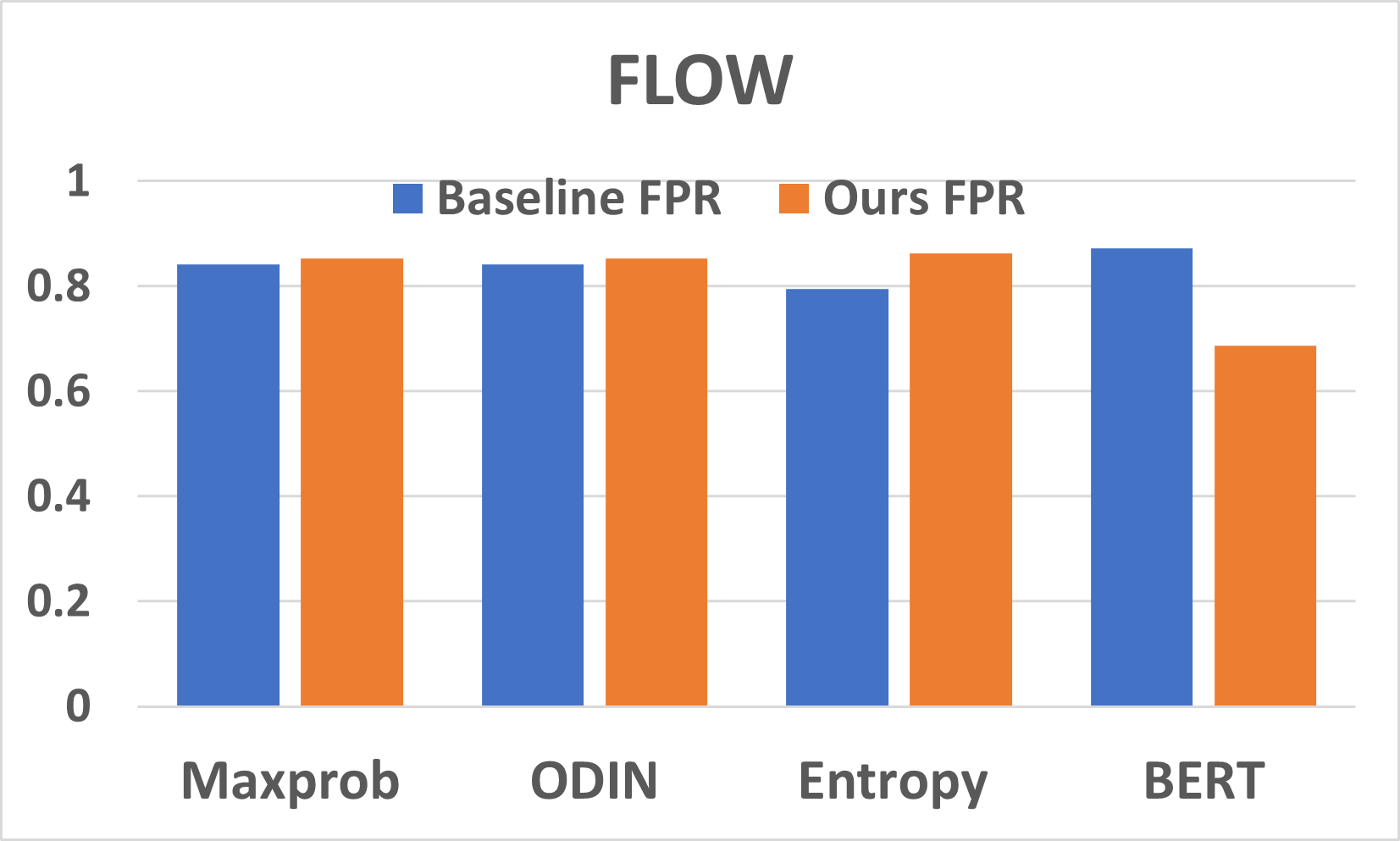}
  \caption{FPR@90 for FLOW}
  \label{fig:sub2b}
\end{subfigure}
%\begin{subfigure}{.9\columnwidth}
%  \centering
%  \includegraphics[width=\linewidth]{EMNLP %2022/figs/rostd.jpg}
%  \caption{AUROC, and FPR for ROSTD}
%  \label{fig:sub3}
%\end{subfigure}
\caption{AUROC ($\uparrow$) and FPR@90 ($\downarrow$) using Maxprob, ODIN, Entropy, and BERT. The average improvements across estimators on AUROC are 5\% on STAR (\ref{fig:sub1a}) and 4\% on FLOW (\ref{fig:sub1b}). The FPR@90 improvements on an average are 8\% on STAR (\ref{fig:sub2a}) and 2\%  on FLOW (\ref{fig:sub2b}).}
\label{fig:auroc}
\end{figure*}

\section{Experiments} \label{exps}
We demonstrate the effectiveness of our proposed approach in this section. For reproducibility of the experiments, we include the codebase in the supplementary. 
%Section \ref{tasks} describe the dataset, Section \ref{setup} talks about the experimental setup, and Section \ref{results} talks about the results.

\subsection{Datasets} \label{tasks}
% To formulate the problem of OOD detection in NLP, it is critical to pick a challenging set of tasks. 
We use three task-oriented dialogue datasets for OOD detection.
% \citet{chen2021gold} used three task-oriented dialogue datasets for OOD detection, 
Namely, Schema-guided Dialog Dataset for Transfer Learning (STAR) \citep{mosig2020star}, SM Calendar flow (FLOW) \citep{andreas2020task}, and Real Out-of-Domain Sentence From Task-oriented Dialog (ROSTD) \citep{gangal2020likelihood}. We follow the data splits and pre-processing steps as described in \citet{chen2021gold}. A detailed description of these tasks is provided in Appendix~\ref{sec:appendix}.

\subsection{Experimental Setup} \label{setup}
Our approach is demonstrated on the BERT pre-trained model \cite{devlin2018bert} with around 110M parameters trained on a single Titan-X GPU. %As explained in Section \ref{methods}, our approach picks keywords that are predicted back similar to the BERT self supervision. 
We optimize $\lambda_{POR}$ for each task through grid search from $1e-5$ to $1e2$ by a factor of 10, and use ${1, 1e-2, 1e-5}$ for STAR, FLOW, and ROSTD respectively. 

This paper compares our approach (\ac{POORE}) with existing baseline OOD detection methods  including Maxprob, Entropy, Mahalanobis, BERT Embed, Gradient Embed, and Dropout. We also consider MASKER~\cite{moon2021masker} as a baseline. A detailed analysis of the differences in performance between all the inference methods and our approach appears in Section \ref{results}. The baseline methods are trained for 25 epochs using the AdamW optimizer with a learning rate of {1e-5, 3e-5, 1e-5} for STAR, FLOW, and ROSTD respectively. We use minimal post hoc fine-tuning of only 1 additional epoch for MASKER and POORE. We use AUROC and FPR@90 to evaluate OOD detection performance. For more details on these metrics, refer to Appendix~\ref{oodmetrics}.

\subsection{Results} \label{results}
Table \ref{mainresults} shows the performance gains from our approach relative to all the baseline methods on three target tasks namely STAR, FLOW, and ROSTD. \ac{POORE} outperforms existing evaluation baselines by significant margins. Specifically on the STAR dataset, relative to Bert Embed and Mahalanobis baselines, we observe 9\% and 4\% respective absolute improvement in AUROC, while observing 26\% and 17\% respective absolute reduction in FPR@90. Similarly on FLOW, the AUROC gains were 13\%, and 1\% relative to BERT Embed and Mahalanobis, while doing worse only on the FPR@90 metric compared to the Mahalanobis baseline. We noted similar consistent gains on the ROSTD.

We also evaluate our framework \ac{POORE} by pairing it with other  confidence estimators like Maxprob, ODIN~\cite{liang2017enhancing}, Entropy, and BERT Embed. Figure \ref{fig:auroc} compares a model trained using \ac{POORE} with a standard model, while using various confidence estimators during inference. As shown in figure~\ref{fig:auroc}, we observe significant gains with our framework over the baseline model for all the confidence estimators. Specifically, the AUROC on FLOW using Bert Embed with \ac{POORE} achieved an improvement of 9\%.  We also pair the above estimators with the MASKER baseline and evaluate these combinations in the ablation shown in Appendix~\ref{maskerbaseline}. In Appendix~\ref{keywordsablation}, we show an ablation comparing our novel keyword selection criterion with the keyword selection criterion used in the MASKER baseline.

\section{Conclusions} \label{conclusion}
In this paper, we propose a novel framework, which we call \ac{POORE}, for improving the robustness of model towards OOD data. Using a combination of Mahalanobis distance and \ac{POR} regularization that maximizes the distance between IND and OOD representations, we demonstrated significant performance gains in a number of target benchmark tasks. Further work could tap into the potential of using external \ac{OOD} data to achieve even more gains over other baselines that use outlier exposure.

% \subsection{References}

% \nocite{Ando2005,borschinger-johnson-2011-particle,andrew2007scalable,rasooli-tetrault-2015,goodman-etal-2016-noise,harper-2014-learning}

% The \LaTeX{} and Bib\TeX{} style files provided roughly follow the American Psychological Association format.
% If your own bib file is named \texttt{custom.bib}, then placing the following before any appendices in your \LaTeX{} file will generate the references section for you:
% \begin{quote}
% \begin{verbatim}
% \bibliographystyle{acl_natbib}
% \bibliography{custom}
% \end{verbatim}
% \end{quote}
% You can obtain the complete ACL Anthology as a Bib\TeX{} file from \url{https://aclweb.org/anthology/anthology.bib.gz}.
% To include both the Anthology and your own .bib file, use the following instead of the above.
% \begin{quote}
% \begin{verbatim}
% \bibliographystyle{acl_natbib}
% \bibliography{anthology,custom}
% \end{verbatim}
% \end{quote}
% Please see Section~\ref{sec:bibtex} for information on preparing Bib\TeX{} files.

\clearpage
\section*{Limitations}

While our work POORE has shown significant gains on the three benchmarks with minimal fine-tuning of a trained classifier, there are a few limitations of our proposed framework. The requirement of pair-wise correspondence for the Euclidean-distance-based regularization in our approach: For OOD data, the other approaches mentioned in related works use KL divergence-based loss, which is not dependent on pair-wise correspondence, but Euclidean-distance-based approaches assume that both vectors used in the distance calculation are in the same space. The approach we employ uses psuedo-OOD data from IND distribution, and hence this is not a limiting factor but may not hold for external OOD data.

\bibliography{arxiv}

\begin{thebibliography}{31}
\expandafter\ifx\csname natexlab\endcsname\relax\def\natexlab#1{#1}\fi

\bibitem[{Aggarwal and Zhai(2012)}]{aggarwal2012survey}
Charu~C Aggarwal and ChengXiang Zhai. 2012.
\newblock A survey of text classification algorithms.
\newblock In \emph{Mining text data}, pages 163--222. Springer.

\bibitem[{Andreas et~al.(2020)Andreas, Bufe, Burkett, Chen, Clausman, Crawford,
  Crim, DeLoach, Dorner, Eisner et~al.}]{andreas2020task}
Jacob Andreas, John Bufe, David Burkett, Charles Chen, Josh Clausman, Jean
  Crawford, Kate Crim, Jordan DeLoach, Leah Dorner, Jason Eisner, et~al. 2020.
\newblock Task-oriented dialogue as dataflow synthesis.
\newblock \emph{Transactions of the Association for Computational Linguistics},
  8:556--571.

\bibitem[{Chen and Yu(2021)}]{chen2021gold}
Derek Chen and Zhou Yu. 2021.
\newblock Gold: improving out-of-scope detection in dialogues using data
  augmentation.
\newblock \emph{arXiv preprint arXiv:2109.03079}.

\bibitem[{Denkowski and Lavie(2011)}]{denkowski2011meteor}
Michael Denkowski and Alon Lavie. 2011.
\newblock Meteor 1.3: Automatic metric for reliable optimization and evaluation
  of machine translation systems.
\newblock In \emph{Proceedings of the sixth workshop on statistical machine
  translation}, pages 85--91.

\bibitem[{Devlin et~al.(2018)Devlin, Chang, Lee, and
  Toutanova}]{devlin2018bert}
Jacob Devlin, Ming-Wei Chang, Kenton Lee, and Kristina Toutanova. 2018.
\newblock Bert: Pre-training of deep bidirectional transformers for language
  understanding.
\newblock \emph{arXiv preprint arXiv:1810.04805}.

\bibitem[{Gal and Ghahramani(2016)}]{gal2016dropout}
Yarin Gal and Zoubin Ghahramani. 2016.
\newblock Dropout as a bayesian approximation: Representing model uncertainty
  in deep learning.
\newblock In \emph{international conference on machine learning}, pages
  1050--1059. PMLR.

\bibitem[{Gangal et~al.(2020)Gangal, Arora, Einolghozati, and
  Gupta}]{gangal2020likelihood}
Varun Gangal, Abhinav Arora, Arash Einolghozati, and Sonal Gupta. 2020.
\newblock Likelihood ratios and generative classifiers for unsupervised
  out-of-domain detection in task oriented dialog.
\newblock In \emph{Proceedings of the AAAI Conference on Artificial
  Intelligence}, volume~34, pages 7764--7771.

\bibitem[{Goodfellow et~al.(2014)Goodfellow, Shlens, and
  Szegedy}]{goodfellow2014explaining}
Ian~J Goodfellow, Jonathon Shlens, and Christian Szegedy. 2014.
\newblock Explaining and harnessing adversarial examples.
\newblock \emph{arXiv preprint arXiv:1412.6572}.

\bibitem[{Hendrycks and Gimpel(2016)}]{hendrycks2016baseline}
Dan Hendrycks and Kevin Gimpel. 2016.
\newblock A baseline for detecting misclassified and out-of-distribution
  examples in neural networks.
\newblock \emph{arXiv preprint arXiv:1610.02136}.

\bibitem[{Hendrycks and Gimpel(2017)}]{hendrycks17baseline}
Dan Hendrycks and Kevin Gimpel. 2017.
\newblock A baseline for detecting misclassified and out-of-distribution
  examples in neural networks.
\newblock \emph{Proceedings of International Conference on Learning
  Representations}.

\bibitem[{Hendrycks et~al.(2018)Hendrycks, Mazeika, and
  Dietterich}]{hendrycks2018deep}
Dan Hendrycks, Mantas Mazeika, and Thomas Dietterich. 2018.
\newblock Deep anomaly detection with outlier exposure.
\newblock \emph{arXiv preprint arXiv:1812.04606}.

\bibitem[{Kamath et~al.(2020)Kamath, Jia, and Liang}]{kamath2020selective}
Amita Kamath, Robin Jia, and Percy Liang. 2020.
\newblock Selective question answering under domain shift.
\newblock \emph{arXiv preprint arXiv:2006.09462}.

\bibitem[{Larson et~al.(2019)Larson, Mahendran, Peper, Clarke, Lee, Hill,
  Kummerfeld, Leach, Laurenzano, Tang, and Mars}]{Larson2019AnED}
Stefan Larson, Anish Mahendran, Joseph Peper, Christopher Clarke, Andrew Lee,
  Parker Hill, Jonathan~K. Kummerfeld, Kevin Leach, Michael Laurenzano, Lingjia
  Tang, and Jason Mars. 2019.
\newblock An evaluation dataset for intent classification and out-of-scope
  prediction.
\newblock \emph{ArXiv}, abs/1909.02027.

\bibitem[{Lee et~al.(2018)Lee, Lee, Lee, and Shin}]{lee2018simple}
Kimin Lee, Kibok Lee, Honglak Lee, and Jinwoo Shin. 2018.
\newblock A simple unified framework for detecting out-of-distribution samples
  and adversarial attacks.
\newblock \emph{Advances in neural information processing systems}, 31.

\bibitem[{Lewis and Gale(1994)}]{lewis1994sequential}
David~D Lewis and William~A Gale. 1994.
\newblock A sequential algorithm for training text classifiers.
\newblock In \emph{SIGIR’94}, pages 3--12. Springer.

\bibitem[{Li et~al.(2017)Li, Monroe, Shi, Jean, Ritter, and
  Jurafsky}]{li2017adversarial}
Jiwei Li, Will Monroe, Tianlin Shi, S{\'e}bastien Jean, Alan Ritter, and Dan
  Jurafsky. 2017.
\newblock Adversarial learning for neural dialogue generation.
\newblock \emph{arXiv preprint arXiv:1701.06547}.

\bibitem[{Li et~al.(2021)Li, Li, Sun, Fan, Zhang, Wu, Meng, and
  Zhang}]{li2021k}
Xiaoya Li, Jiwei Li, Xiaofei Sun, Chun Fan, Tianwei Zhang, Fei Wu, Yuxian Meng,
  and Jun Zhang. 2021.
\newblock $ k $ folden: $ k $-fold ensemble for out-of-distribution detection.
\newblock \emph{arXiv preprint arXiv:2108.12731}.

\bibitem[{Liang et~al.(2017)Liang, Li, and Srikant}]{liang2017enhancing}
Shiyu Liang, Yixuan Li, and Rayadurgam Srikant. 2017.
\newblock Enhancing the reliability of out-of-distribution image detection in
  neural networks.
\newblock \emph{arXiv preprint arXiv:1706.02690}.

\bibitem[{Moon et~al.(2021)Moon, Mo, Lee, Lee, and Shin}]{moon2021masker}
Seung~Jun Moon, Sangwoo Mo, Kimin Lee, Jaeho Lee, and Jinwoo Shin. 2021.
\newblock Masker: Masked keyword regularization for reliable text
  classification.
\newblock In \emph{Proceedings of the AAAI Conference on Artificial
  Intelligence}, volume~35, pages 13578--13586.

\bibitem[{Mosig et~al.(2020)Mosig, Mehri, and Kober}]{mosig2020star}
Johannes~EM Mosig, Shikib Mehri, and Thomas Kober. 2020.
\newblock Star: A schema-guided dialog dataset for transfer learning.
\newblock \emph{arXiv preprint arXiv:2010.11853}.

\bibitem[{Nalisnick et~al.(2019)Nalisnick, Matsukawa, Teh, and
  Lakshminarayanan}]{Nalisnick2019DetectingOI}
Eric~T. Nalisnick, Akihiro Matsukawa, Yee~Whye Teh, and Balaji
  Lakshminarayanan. 2019.
\newblock Detecting out-of-distribution inputs to deep generative models using
  a test for typicality.
\newblock \emph{ArXiv}, abs/1906.02994.

\bibitem[{Podolskiy et~al.(2021)Podolskiy, Lipin, Bout, Artemova, and
  Piontkovskaya}]{podolskiy2021revisiting}
Alexander Podolskiy, Dmitry Lipin, Andrey Bout, Ekaterina Artemova, and Irina
  Piontkovskaya. 2021.
\newblock Revisiting mahalanobis distance for transformer-based out-of-domain
  detection.
\newblock In \emph{Proceedings of the AAAI Conference on Artificial
  Intelligence}, volume~35, pages 13675--13682.

\bibitem[{Ren et~al.(2019)Ren, Liu, Fertig, Snoek, Poplin, DePristo, Dillon,
  and Lakshminarayanan}]{ren2019likelihood}
Jie Ren, Peter~J. Liu, Emily Fertig, Jasper Snoek, Ryan Poplin, Mark~A.
  DePristo, Joshua~V. Dillon, and Balaji Lakshminarayanan. 2019.
\newblock \emph{Likelihood Ratios for Out-of-Distribution Detection}. Curran
  Associates Inc., Red Hook, NY, USA.

\bibitem[{Shen et~al.(2019)Shen, Cheng, Sundararaman, Zhang, Yang, Tang,
  Celikyilmaz, and Carin}]{shen2019learning}
Dinghan Shen, Pengyu Cheng, Dhanasekar Sundararaman, Xinyuan Zhang, Qian Yang,
  Meng Tang, Asli Celikyilmaz, and Lawrence Carin. 2019.
\newblock Learning compressed sentence representations for on-device text
  processing.
\newblock \emph{arXiv preprint arXiv:1906.08340}.

\bibitem[{Siedlikowski et~al.(2021)Siedlikowski, No{\"e}l, Moynihan, Robin
  et~al.}]{siedlikowski2021chloe}
Sophia Siedlikowski, Louis-Philippe No{\"e}l, Stephanie~Anne Moynihan, Marc
  Robin, et~al. 2021.
\newblock Chloe for covid-19: Evolution of an intelligent conversational agent
  to address infodemic management needs during the covid-19 pandemic.
\newblock \emph{Journal of Medical Internet Research}, 23(9):e27283.

\bibitem[{Sundararaman et~al.(2020)Sundararaman, Si, Subramanian, Wang,
  Hazarika, and Carin}]{sundararaman2020methods}
Dhanasekar Sundararaman, Shijing Si, Vivek Subramanian, Guoyin Wang, Devamanyu
  Hazarika, and Lawrence Carin. 2020.
\newblock Methods for numeracy-preserving word embeddings.
\newblock In \emph{Proceedings of the 2020 Conference on Empirical Methods in
  Natural Language Processing (EMNLP)}, pages 4742--4753.

\bibitem[{Sundararaman et~al.(2019)Sundararaman, Subramanian, Wang, Si, Shen,
  Wang, and Carin}]{sundararaman2019syntax}
Dhanasekar Sundararaman, Vivek Subramanian, Guoyin Wang, Shijing Si, Dinghan
  Shen, Dong Wang, and Lawrence Carin. 2019.
\newblock Syntax-infused transformer and bert models for machine translation
  and natural language understanding.
\newblock \emph{arXiv preprint arXiv:1911.06156}.

\bibitem[{Sundararaman et~al.(2022)Sundararaman, Subramanian, Wang, Xu, and
  Carin}]{sundararaman2022number}
Dhanasekar Sundararaman, Vivek Subramanian, Guoyin Wang, Liyan Xu, and Lawrence
  Carin. 2022.
\newblock Number entity recognition.
\newblock \emph{arXiv preprint arXiv:2205.03559}.

\bibitem[{Vaswani et~al.(2017)Vaswani, Shazeer, Parmar, Uszkoreit, Jones,
  Gomez, Kaiser, and Polosukhin}]{vaswani2017attention}
Ashish Vaswani, Noam Shazeer, Niki Parmar, Jakob Uszkoreit, Llion Jones,
  Aidan~N Gomez, {\L}ukasz Kaiser, and Illia Polosukhin. 2017.
\newblock Attention is all you need.
\newblock \emph{Advances in neural information processing systems}, 30.

\bibitem[{Yang et~al.(2021)Yang, Zhou, Li, and Liu}]{yang2021generalized}
Jingkang Yang, Kaiyang Zhou, Yixuan Li, and Ziwei Liu. 2021.
\newblock Generalized out-of-distribution detection: A survey.
\newblock \emph{arXiv preprint arXiv:2110.11334}.

\bibitem[{Zheng et~al.(2020)Zheng, Chen, and Huang}]{zheng2020OODDialog}
Yinhe Zheng, Guanyi Chen, and Minlie Huang. 2020.
\newblock \href {https://doi.org/10.1109/TASLP.2020.2983593} {Out-of-domain
  detection for natural language understanding in dialog systems}.
\newblock \emph{IEEE/ACM Trans. Audio, Speech and Lang. Proc.}, 28:1198–1209.

\end{thebibliography}
\bibliographystyle{acl_natbib}

\begin{acronym}[CCCCCCCC]
    \acro{OOD}{Out-of-Distribution}
    \acro{IND}{In-Distribution}
    \acro{POORE}{POsthoc pseudo Ood REgularization}
    \acro{POR}{Pseudo Ood Regularization}
\end{acronym}

\clearpage
\newpage
\appendix
\section{Tasks}
\label{sec:appendix}
\noindent \textbf{STAR.} This is a dialog dataset with 6651 dialogues spanning multiple domains and intents \cite{mosig2020star}. Responses to dialogs that were marked either “ambiguous” or “out-of-scope” are used as OOD examples. The dataset has 29,104 examples with 104 intent labels. \\

\noindent \textbf{FLOW.} The FLOW dataset is a semantic parsing dataset with annotations for each turn of a dialog \cite{andreas2020task}. In FLOW, the OOD samples are from discussions where the user stays far away from the central topic. The dataset has 71,551 examples spanning 44 intents. \\

\noindent \textbf{ROSTD.} \citet{gangal2020likelihood} designed ROSTD, a dataset proposed for OOD detection. They use external source for OOD samples, while the internal data represents IND. ROSTD contains 47,913 examples with 13 classes. 

\section{OOD Evaluation Metrics}
\label{oodmetrics}
\noindent \ac{OOD} detection is evaluated using the Area Under the Receiver-Operating Curve (AUROC) metric for the binary classification task based on the estimated confidence score. An \ac{OOD} method that perfectly separates $s(x^{ind})$ from $s(x^{ood})$ achieves an AUROC score of 100\%. Another common metric used to evaluate \ac{OOD} detection is false positive rate (FPR) at a fixed recall.
% This is a section in the appendix.

\section{Adaptation of MASKER Baseline}
\label{maskerbaseline}

In the table \ref{adaptedresults}, improvised Masker baseline results can be seen, which include the results on a number of evaluation metrics. While the performance using improvised baseline is better than the GOLD baseline, our approach beats this model considerably.

\begin{table}[!th]
\centering
\resizebox{\columnwidth}{!}{%
\begin{tabular}{l|c|c|c|c|c|c}
\toprule
Methods & \multicolumn{2}{c}{STAR} & \multicolumn{2}{|c}{FLOW}  & \multicolumn{2}{|c}{ROSTD} \\ \midrule
& AUROC $\uparrow$ & FPR@90 $\downarrow$  & AUROC $\uparrow$ & FPR@90 $\downarrow$ & AUROC $\uparrow$ & FPR@90 $\downarrow$ \\ \midrule
Maxprob     & 71.54 &  7282   & 68.16  & 67.52   & 86.95 & 54.26\\ 
ODIN        & 72.45 &  71.86   & 68.57  &  66.81   & 86.86 &  54.25\\
BERT        & 75.93    & 62.89   & 69.79    & 70.07   & 99.16 &  1.73\\
Mahalanobis & 79.38    & 59.97   & 72.99 & 65.23   & 99.41 &  1.15\\
POORE (Ours)     & 81.11 & 48.30 & 74.08 & 69.26 & 99.51 & 0.97 \\
\bottomrule
\end{tabular}%
}
\caption{Masker baseline and Adapted approaches.}
\label{adaptedresults}
\end{table}

\newpage
\section{Ablation for choosing keywords}
\label{keywordsablation}

Table \ref{keywordstable} compares the \ac{OOD} detection performance of our proposed keyword selection approach described in Section~\ref{maskingood} with the keyword selection criterion in the baseline MASKER in our proposed ~\ac{POORE} framework.

\begin{table}
\small
\centering
\resizebox{0.55\textwidth}{!}{%
\begin{tabular}{l||c|c|c|c|c|c}
\toprule
Methods & \multicolumn{2}{c}{STAR} & \multicolumn{2}{|c}{FLOW}  & \multicolumn{2}{|c}{ROSTD} \\ \midrule
& AUROC $\uparrow$ & FPR@90 $\downarrow$  & AUROC $\uparrow$ & FPR@90 $\downarrow$ & AUROC $\uparrow$ & FPR@90 $\downarrow$ \\ \midrule
POORE with baseline keywords & 80.69    & 58.71   & 73.41 & 68.08   & 99.52 &  1.06\\ %
POORE (Ours)     & 81.11 & 48.30 & 74.08 & 69.26 & 99.51 & 0.97    \\ % \hline
\bottomrule
\end{tabular}%
}
\caption{Ablation for keywords}
\label{keywordstable}
\end{table}

\end{document}